\documentclass[runningheads]{llncs}
\usepackage{graphicx}

\usepackage{tikz}
\usepackage{comment} 
\usepackage{amsmath,amssymb} 
\usepackage{color}

\usepackage{multirow}
\usepackage{adjustbox}
\usepackage{textgreek}
\usepackage{amsmath}
\usepackage{bm}
\usepackage{algorithm}
\usepackage{algorithmic}
\usepackage{graphicx}
\usepackage{caption}

\begin{document}
\pagestyle{headings}
\mainmatter
\def\ECCVSubNumber{3705}  

\title{Improving 3D Object Detection through Progressive Population Based Augmentation} 


\titlerunning{Progressive Population Based Augmentation}
\author{
Shuyang Cheng\inst{1} \and
Zhaoqi Leng\thanks{Work done while at Google LLC.}\inst{1} \and
Ekin Dogus Cubuk\inst{2} \and
Barret Zoph\inst{2} \and
Chunyan Bai \inst{1} \and
Jiquan Ngiam \inst{2} \and
Yang Song \inst{1} \and
Benjamin Caine \inst{2} \and
Vijay Vasudevan \inst{2} \and
Congcong Li \inst{1} \and
Quoc V. Le \inst{2} \and
Jonathon Shlens \inst{2} \and
Dragomir Anguelov \inst{1}
}
\authorrunning{S. Cheng et al.}
\institute{Waymo LLC \and Google LLC}
\maketitle

\begin{abstract}
Data augmentation has been widely adopted for object detection in 3D point clouds. However, all previous related efforts have focused on manually designing specific data augmentation methods for individual architectures. In this work, we present the first attempt to automate the design of data augmentation policies for 3D object detection. We introduce the Progressive Population Based Augmentation (PPBA) algorithm, which learns to optimize augmentation strategies by narrowing down the search space and adopting the best parameters discovered in previous iterations. On the KITTI 3D detection test set, PPBA improves the StarNet detector by substantial margins on the moderate difficulty category of cars, pedestrians, and cyclists, outperforming all current state-of-the-art single-stage detection models. Additional experiments on the Waymo Open Dataset indicate that PPBA continues to effectively improve the StarNet and PointPillars detectors on a 20x larger dataset compared to KITTI. The magnitude of the improvements may be comparable to advances in 3D perception architectures and the gains come without an incurred cost at inference time. In subsequent experiments, we find that PPBA may be up to 10x more data efficient than baseline 3D detection models without augmentation, highlighting that 3D detection models may achieve competitive accuracy with far fewer labeled examples.

\end{abstract}
\section{Introduction}

LiDAR is a prominent sensor for autonomous driving and robotics because it provides detailed 3D information critical for perceiving and tracking real-world objects \cite{cho2014multi,thrun2006stanley}.
The 3D localization of objects within LiDAR point clouds represents one of the most important tasks in visual perception, and much effort has focused on developing novel network architectures for point clouds \cite{chen2017multi,yang2018pixor,luo2018fast,yang2018hdnet,zhou2018voxelnet,yan2018second,lang2019pointpillars,shi2019pv}. Following the image classification literature, such modeling efforts have employed manually designed data augmentation schemes for boosting performance \cite{chen2017multi,yan2018second,lang2019pointpillars,yang2018pixor,ngiam2019starnet,zhou2019end,shi2019pv,zhou2018voxelnet}.

In recent years, much work in the 2D image literature has demonstrated that investing heavily into data augmentation may lead to gains comparable to those obtained by advances in model architectures~\cite{cubuk2018autoaugment,zoph2019learning,lim2019fast,ho2019population,cubuk2019randaugment}. Despite this, 3D detection models have yet to significantly leverage automated data augmentation methods (but see \cite{li2020pointaugment}).
Naively porting ideas that are effective for images to point cloud data presents numerous challenges, as the the types of augmentations appropriate for point clouds differ tremendously. Transformations appropriate for point clouds are typically geometric-based and may contain a large number of parameters. Thus, the search space proposed in \cite{cubuk2018autoaugment,zoph2019learning} may not be naively reused for an automated search in point cloud augmentation space. Finally, because the search space is far larger, employing a more efficient search method becomes a practical necessity.
Several works have attempted to significantly accelerate the search for data augmentation strategies \cite{lim2019fast,ho2019population,cubuk2019randaugment}, however it is unclear if such methods transfer successfully to point clouds.

In this work, we demonstrate that automated data augmentation significantly improves the prediction accuracy of 3D object detection models. We introduce a new search space for point cloud augmentations in 3D object detection. In this search space, we find the performance distribution of augmentation policies is quite diverse. To effectively discover good augmentation policies, we present an evolutionary search algorithm termed {\it Progressive Population Based Augmentation} (PPBA). PPBA works by narrowing down the search space through successive iterations of evolutionary search, and by adopting the best parameters discovered in past iterations. We demonstrate that PPBA is effective at finding good data augmentation strategies across datasets and detection architectures. Additionally, we find that a model trained with PPBA may be up to 10x more data efficient, implying reduced human labeling demands for point clouds.

Our main contributions can be summarized as follows: (1) We propose an automated data augmentation technique for localization in 3D point clouds. (2) We demonstrate that the proposed search method effectively improves point cloud 3D detection models compared to random search with less computational cost. (3) We demonstrate up to a 10x increase in data efficiency when employing PPBA. (4) Beyond 3D detection, we also demonstrate that PPBA generalizes to 2D image classification.
\section{Related Work}
Data augmentation has been an essential technique for boosting the performance of 2D image classification and object detection models. Augmentation methods typically include manually designed image transformations, to which the labels remain invariant, or distortions of the information present in the images. For example, elastic distortions, scale transformations, translations, and rotations are beneficial on models trained on MNIST~\cite{simard2003best,ciregan2012multi,wan2013regularization,sato2015apac}. Crops, image mirroring and color shifting / whitening~\cite{krizhevsky2012imagenet} are commonly adopted on natural image datasets like CIFAR-10 and ImageNet. Recently, cutout~\cite{cutout2017} and mixup~\cite{zhang2017mixup} have emerged as data augmentation methods that lead to good improvements in natural image datasets. For object detection in 2D images, image mirroring and multi-scale training are popular distortions~\cite{girshick2018detectron}. Dwibedi et al. add new objects on training images by cut-and-paste~\cite{dwibedi2017cut}.

While the augmentation operations mentioned above are designed by domain experts, there are also automated approaches to designing data augmentation strategies for 2D images. Early attempts include Smart Augmentation, which uses a network to generate augmented data by merging two or more image samples~\cite{lemley2017smart}. Ratner et al. use GANs to output sequences of data augmentation operations~\cite{ratner2017learning}. AutoAugment uses reinforcement learning to optimize data augmentation strategies for classification~\cite{cubuk2018autoaugment} and object detection~\cite{zoph2019learning}. More recently, improved search methods are able to find data augmentation strategies more efficiently~\cite{cubuk2019randaugment,ho2019population,lim2019fast}.

While all the mentioned work so far is on 2D image classification and object detection, automated data augmentation methods have not been explored for 3D object detection tasks to the best of our knowledge. Models trained on KITTI use a wide variety of manually designed distortions. Due to the small size of the KITTI training set, data augmentation has been shown to improve performance significantly (common augmentations include horizontal flips, global scale distortions, and rotations)~\cite{chen2017multi,yan2018second,yang2018pixor,lang2019pointpillars,shi2019pv}. Yan et al. add new objects in training point clouds by pasting points inside ground truth 3D bounding boxes~\cite{yan2018second}. Despite its effectiveness for KITTI models, data augmentation was not used on some of the larger point cloud datasets~\cite{ngiam2019starnet,zhou2019end}. Very recently, an automated data augmentation approach was studied for point cloud classification~\cite{li2020pointaugment}.

Historically, 2D vision research has focused on architectural modifications to improve generalization. More recently, it was observed that improving data augmentation strategies can lead to comparable gains to a typical architectural advance~\cite{zhang2017mixup,fang2019instaboost,cubuk2018autoaugment,zoph2019learning}. In this work, we demonstrate that a similar type of improvement can also be obtained by an effective automated data augmentation strategy for 3D object detection over point clouds.
\section{Methods}
We formulate the problem of finding the right augmentation strategy as a special case of hyperparameter schedule learning. The proposed method consists of two components: a specialized data augmentation search space for point cloud inputs and a search algorithm for the optimization of data augmentation parameters. We describe these two components below.

\subsection{Search Space for 3D Point Cloud Augmentation}\label{sec:search_space}
In the proposed search space, an augmentation policy consists of N augmentation operations. Additionally, each operation is associated with a probability and some specialized parameters. For example, the ground-truth augmentation operation has parameters denoting the probability for sampling vehicles, pedestrians, cyclists, etc.; the global translation noise operation has parameters for the distortion magnitude of the translation operation on x, y and z coordinates. To reduce the size of the search space and increase the diversity of the training data, these different operations are always applied according to some learned probabilities in the same, pre-determined order to point clouds during training.

\begin{figure}[t]
    \centering
    \includegraphics[width=1\linewidth]{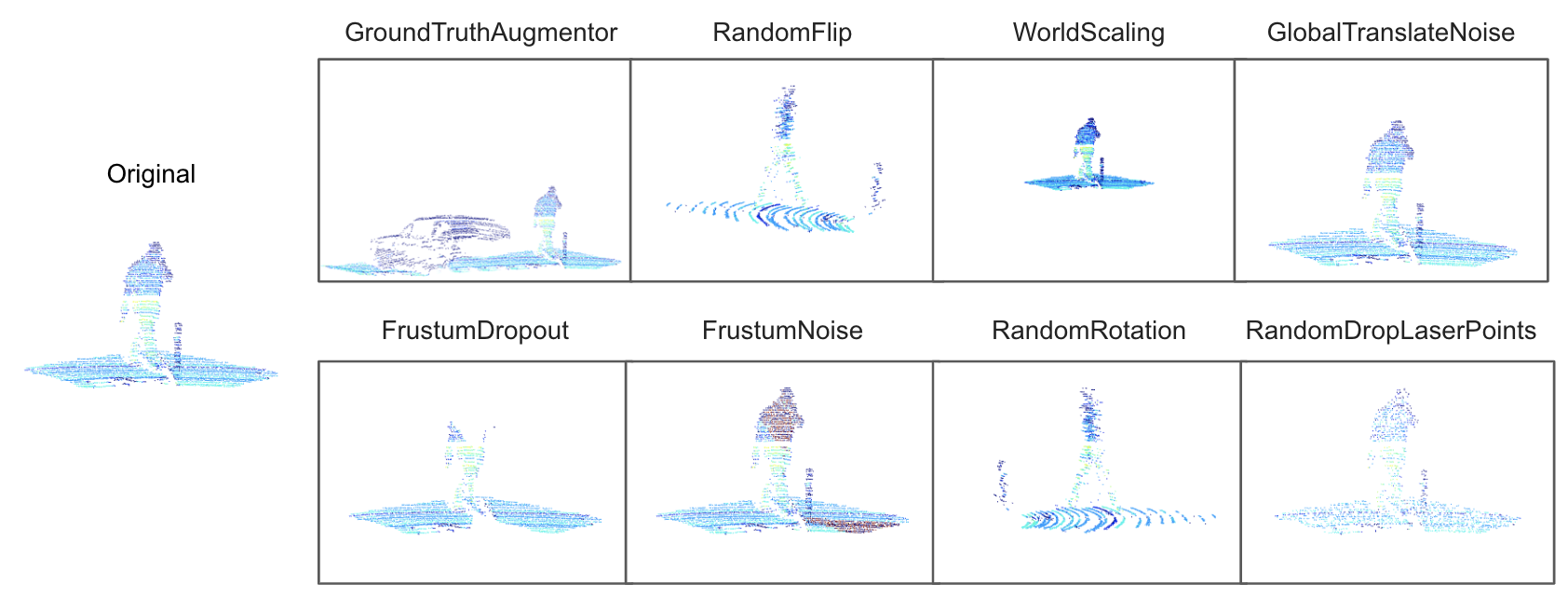}
    \caption{\textbf{Visualization of the augmentation operations in the proposed search space.} An augmentation policy is defined by a list of distinct augmentation operations and the corresponding augmentation parameters. Details of these operations are in Table 7 and Table 8 in the Appendix. }
    \label{fig:augmentation_operations}
\end{figure}

The basic augmentation operations in the proposed search space fall into two main categories: {\it global operations}, which are applied to all points in a frame (such as rotation along Z-axis, coordinate scaling, etc.), and {\it local operations}, which are applied to points locally (such as dropping out points within a frustum, pasting points within bounding boxes from other frames, etc.). Our list of augmentation operations (see Fig.~\ref{fig:augmentation_operations}) consists of GroundTruthAugmentor, RandomFlip, WorldScaling, GlobalTranslateNoise, FrustumDropout, FrustumNoise, RandomRotation and RandomDropLaserPoints. In total, there are 8 augmentation operations and 29 operation parameters in the proposed search space.

\subsection{Learning through Progressive Population Based Search}
The proposed search process is maximizing a given metric {$\Omega$} on a model {$\theta$} by optimizing a schedule of augmentation operation parameters {$\bm{\lambda}=(\lambda_t)^T_{t=1}$}, where $t$ represents the number of iterative updates for the augmentation operation parameters during model training. For point cloud detection tasks, we use mean average precision (mAP) as the performance metric. The search process for the best augmentation schedule $\bm{\lambda^*}$ optimizes:
\begin{align}
\bm{\lambda^*} = \arg\max_{\bm{\lambda} \in \Lambda^T} \Omega(\theta)
\label{eq:1}
\end{align}

During training, the objective function {\it L} (which is used for optimization of the model  {$\theta$} given data and label pairs $(\mathcal{X}, \mathcal{Y})$) is usually different from the actual performance metric {$\Omega$}, since the optimization procedure (i.e. stochastic gradient descent) requires a differentiable objective function. Therefore, at each iteration $t$ the model {$\theta$} is optimizing:
\begin{align}
\bm{\theta^*_t} = \arg\min_{\theta \in \Theta} L(\mathcal{X}, \mathcal{Y}, \lambda^t)
\label{eq:2}
\end{align}

\begin{figure}[h!]
    \centering
    \includegraphics[width=\linewidth]{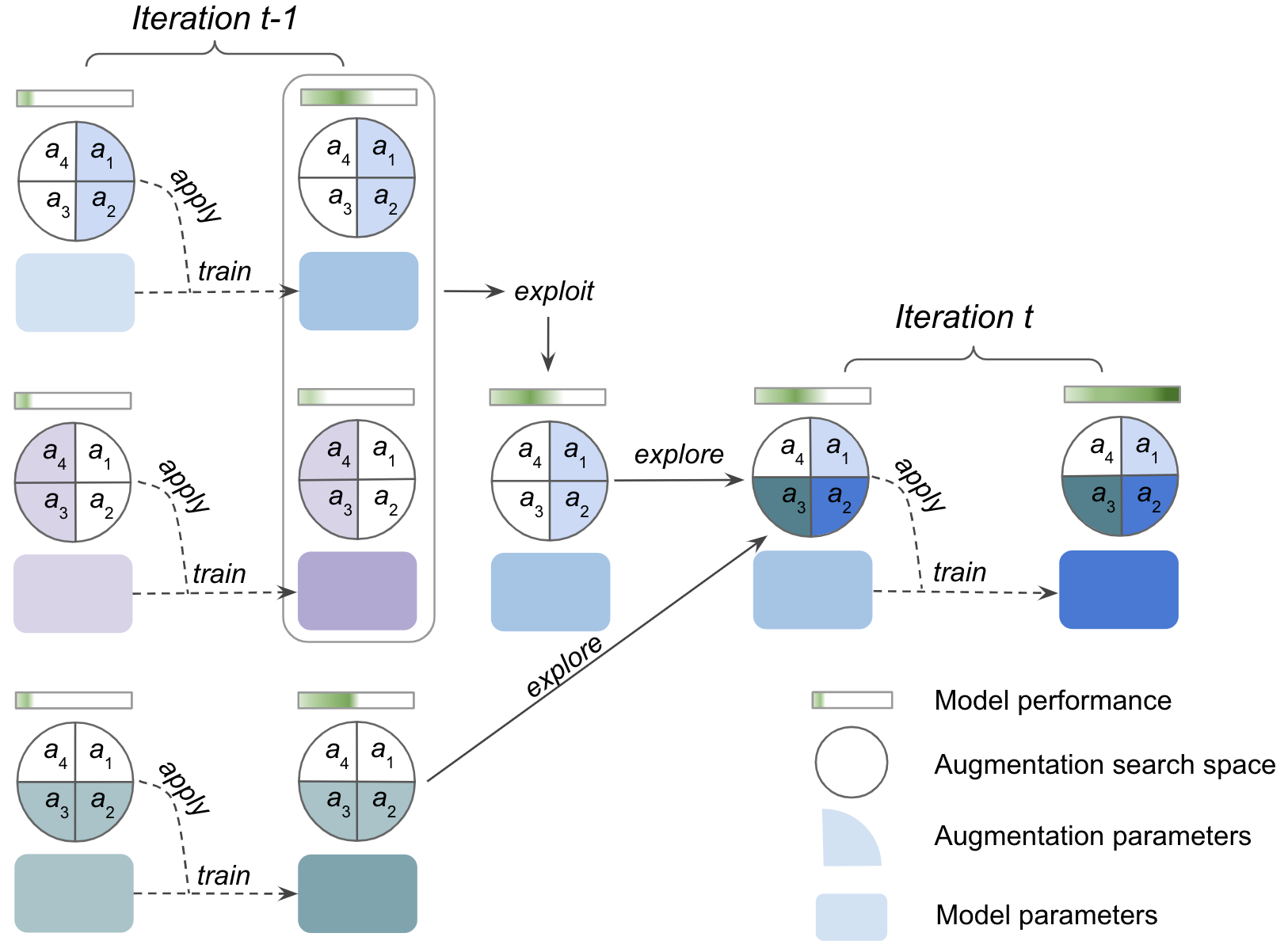}
    \caption{\textbf{An example of Progressive Population Based Augmentation.} Four augmentation operations ($a_1$, $a_2$, $a_3$, $a_4$) are applied to the input data during training; their parameter set comprises the full search space. During progressive population based search, only two augmentation operations out of the four are explored for optimization at every iteration. For example, at the beginning of iteration $t-1$, augmentation parameters of ($a_1$, $a_2$) are selected for exploration for the blue model while augmentation parameters of ($a_3$, $a_4$) are selected for exploration for the purple model. At the end of training in iteration $t-1$, an inferior model (the purple model) is \textit{exploited} by a model with higher performance (the blue model). Afterwards, a successor will inherit both model parameters and augmentation parameters from the winner model - the blue model. During the \textit{exploration} phase, the selected augmentation operations for exploration by the successor model are randomly sampled and become ($a_2$, $a_3$). Since augmentation parameters of $a_3$ have not been explored by the predecessor (the blue model), corresponding augmentation parameters of the best model (the green model), in which $a_3$ has been selected for exploration, will be adopted for exploration by the successor model.}\label{fig:ppba_algorithm}
\end{figure}

During search, the training process of the model is split into $\mathcal{N}$ iterations. At every iteration, $\mathcal{M}$ models with different $\lambda_t$ are trained in parallel and are afterwards evaluated with the metric {$\Omega$}. Models trained in all previous iterations are placed in a population $\mathcal{P}$. In the initial iteration, all model parameters and augmentation parameters are randomly initialized. After the first iteration, model parameters are determined through an {\it exploit} phase - inheriting from a better performing parent model by exploiting the rest of the population $\mathcal{P}$. The {\it exploit} phase is followed by an {\it exploration} phase, in which a subset of the augmentation operations will be explored for optimization by mutating the corresponding augmentation parameters in the parent model, while the remaining augmentation parameters will be directly inherited from the parent model.

Similar to Population Based Training~\cite{pbt2017}, the {\it exploit} phase will keep the good models and replace the inferior models at the end of every iteration. In contrast with Population Based Training, the proposed method focuses only on a subset of the search space at each iteration. During the {\it exploration} phase, a successor might focus on a different subset of the parameters than its predecessor. In that case, the remaining parameters (parameters that the predecessor does not focus on) are mutated based on the parameters of the corresponding operations with the best overall performance. In Fig~\ref{fig:ppba_algorithm}, we show an example of Progressive Population Based Augmentation. The complete PPBA algorithm is described in detail in Algorithm 1 in the Appendix. 

\subsection{Schedule Optimization with Historical Data}
The search spaces for data augmentation are different between 2D images and 3D point clouds. For example, each operation in the AutoAugment~\cite{cubuk2018autoaugment} search space for 2D images has a single parameter. Furthermore, any value of this parameter within the predefined range leads to a reasonable image. For this reason, even sampling random augmentation policies for 2D images leads to some improvement in generalization~\cite{cubuk2018autoaugment,cubuk2019randaugment}. On the other hand, the augmentation operations for 3D point clouds are much harder to optimize. Each operation has several parameters, and a good range for these parameters is not known a priori. For example, there are five parameters -- theta\_width, phi\_width, distance, keep\_prob and drop\_type -- in the FrustumDropout operation. The analogous operation for 2D images is cutout~\cite{cutout2017}, which has only one parameter. Therefore it is more challenging to discover optimal parameters for point cloud operations with limited resources.

In order to learn the parameters for individual operations effectively, PPBA modifies only a small portion of the parameters in the search space at every iteration, and the historical information from the previous iterations are reused to optimize the augmentation schedule. By narrowing down the focus on certain subsets of the search space, it becomes easier to distinguish the inferior augmentation parameters. To mitigate the slowing down of search speed caused by the search space shrinkage at each training iteration, the best parameters of each operation discovered in the past iterations are adopted by the successors, when their focused subsets of the search space are different from their predecessors.

\begin{figure}[h]
    \centering
    \includegraphics[width=\linewidth]{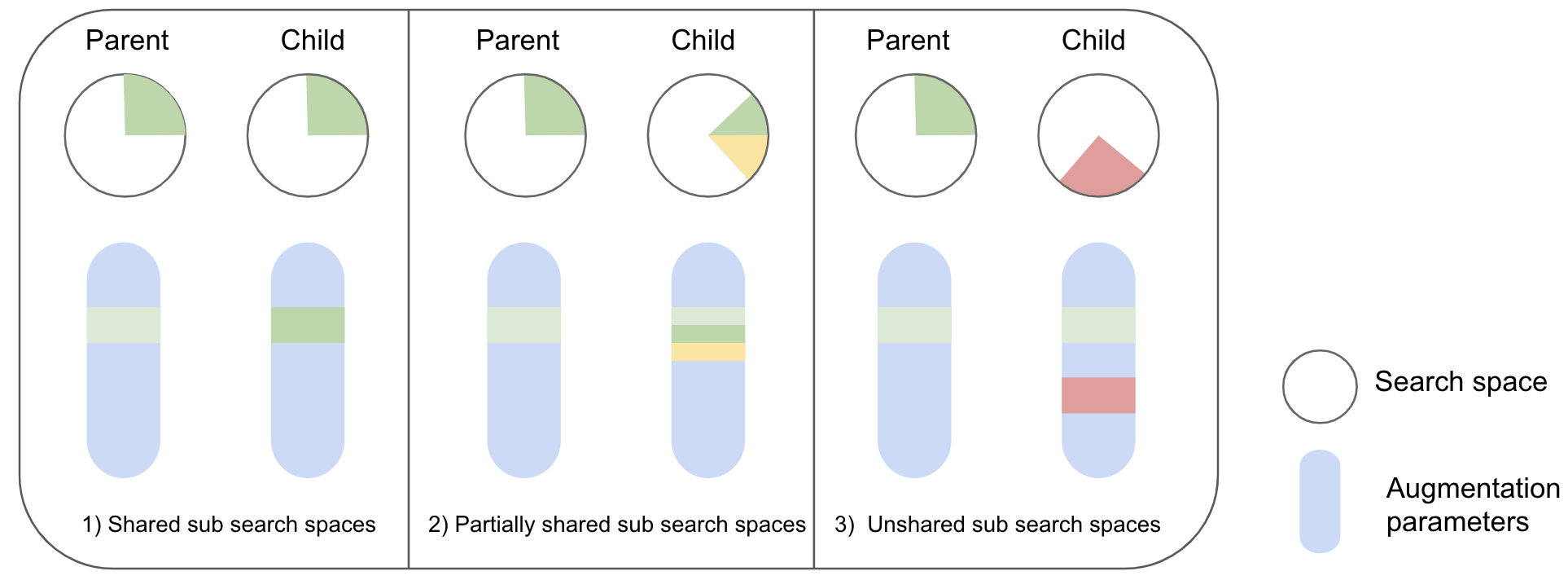}
    \caption{\textbf{Three types of scenarios for the subsets of the search space explored by the parent and the child models.} 1) The subsets are the same. 2) The subsets are partially shared. 3) The subsets are unshared. In both 1) and 2), the overlapped augmentation parameters for exploration in the child model are mutated based on the corresponding parameters in the parent model (updating from light green to dark green). In both 2) and 3), the non-overlapped augmentation parameters for exploration in the child model are mutated based on the best augmentation parameters discovered in the past iterations (if available) or random sampling (updating from blue to yellow/red).}
    \label{fig:search_space_exploration}
\end{figure}

In Fig~\ref{fig:search_space_exploration}, we show three types of scenarios for the subsets of the search space explored by a successor and its predecessor. The details of the exploration phase based on historical data are described in Algorithm 2 in the Appendix.

\section{Experiments}
In this section, we empirically investigate the performance of PPBA on predictive accuracy, computational efficiency and data efficiency.
We focus on single-stage detection models due to their simplicity, speed advantages and widespread adoption~\cite{yan2018second,lang2019pointpillars,ngiam2019starnet}.

We first benchmark PPBA on the KITTI object detection benchmark~\cite{kitti2013} and the Waymo Open Dataset~\cite{waymo2019} (Sections \ref{sec:kitti_result} and \ref{sec:waymo_result}). Our results show PPBA improves the baseline models and the improvement magnitude may be comparable to advances in 3D perception architectures. Next, we compare PPBA with random search and PBA~\cite{ho2019population} on the KITTI Dataset (Section \ref{sec:computation_efficiency}) to demonstrate PPBA's effectiveness and efficiency. In addition, we study the data efficiency of PPBA on the Waymo Open Dataset (Section~\ref{sec:data_efficiency}). Our experiments show that PPBA can achieve competitive accuracy with far fewer labeled examples compared with no augmentation. Finally, the PPBA algorithm was designed for, but is not limited to, 3D object detection tasks. We study its ability to generalize to 2D image classification and present results in Section~\ref{sec:ppba_generalization}. 

\subsection{Surpassing Single-Stage Models on the KITTI Dataset}\label{sec:kitti_result}
The KITTI Dataset~\cite{kitti2013} is generally recognized to be a small dataset for modern methods, and thus, data augmentation is critical to the performance of models trained on it~\cite{yan2018second,lang2019pointpillars,ngiam2019starnet}.
We evaluate PPBA with StarNet~\cite{ngiam2019starnet} on the KITTI {\it test} split in Table~\ref{table:kitti_test}. PPBA improves the detection performance of StarNet significantly, outperforming all current state-of-the-art single-stage point cloud detection models on the moderate difficulty category.

\begin{table}[h!]
\small{
\begin{center}
\caption{Performance comparison of single-stage point cloud detection models on the KITTI {\it test} set using 3D evaluation. mAP is calculated with an IOU of 0.7, 0.5 and 0.5 for vehicles,
cyclists and pedestrians, respectively}
\label{table:kitti_test}
\begin{tabular}{l||ccc|ccc|ccc}
\hline
\multirow{2}{*}{Method} & 
\multicolumn{3}{c|}{Car} & \multicolumn{3}{c|}{Pedestrian} & \multicolumn{3}{c}{Cyclist} \\
& Easy & Mod. & Hard & Easy & Mod. & Hard & Easy & Mod. & Hard\\
\hline
ContFuse \cite{contfuse2018} & 83.68 &  68.78 & 61.67 & - & - & - & - & - & - \\
VoxelNet \cite{zhou2018voxelnet} & 77.47 & 65.11 & 57.73 & 39.48 & 33.69 & 31.51 & 61.22 & 48.36 & 44.37 \\
SECOND \cite{yan2018second} &  83.13 & 73.66 & 66.20 & 51.07 & 42.56 & 37.29 & 70.51 & 53.85 & 46.90 \\
3D IoU Loss \cite{iouloss2019} & \bf84.43 & 76.28 & 68.22 & - & - & - & - & - & - \\
PointPillars \cite{lang2019pointpillars} & 79.05 & 74.99 & 68.30 & 52.08 & 43.53 & 41.49 & 75.78 & 59.07 & 52.92 \\
\hline
StarNet~\cite{ngiam2019starnet} & 81.63 & 73.99 & 67.07 & 48.58 & 41.25 & 39.66 & 73.14 & 58.29 & 52.58 \\
StarNet~\cite{ngiam2019starnet} + PPBA & 84.16 & \bf77.65 & \bf71.21 & \bf52.65 & \bf44.08 & \bf41.54 & \bf79.42 & \bf61.99 & \bf55.34 \\
\hline
\end{tabular}
\end{center}
}
\end{table}

During the PPBA search, 16 trials are trained to optimize the mAP for car (30 iterations) and for pedestrian/cyclist (20 iterations), respectively. The same training and inference settings\footnote{\url{http://github.com/tensorflow/lingvo}} as~\cite{ngiam2019starnet} are used, while all trials are trained on the {\it train} split (3,712 samples) and validated on the {\it validation} split (3,769 samples). We train the first iteration for 3,000 steps, and all subsequent iterations for 1,000 steps with batch size 64. The search is conducted in the search space described in Section~\ref{sec:search_space}.

Manually designed augmentation policies are typically kept constant during training. In contrast, stochasticity lies at the heart of the augmentation policies in PPBA, i.e. each operation is applied stochastically and its parameters evolve as the training progresses. We have found that simply using the final parameters discovered by PPBA gets worse results than PPBA.

We use GroundTruthAugmentor to highlight the difference between the manually designed and learned augmentation policies. While training a StarNet vehicle detection model on KITTI with PPBA, the probability of applying the operation decreases from 100\% to 16\% and the probability of pasting vehicles reduces from 100\% to 21\%, while the probability of pasting pedestrians and cyclists increases from 0\% to 28\% and 8\% respectively. This suggests that pasting the object of interest in every frame during training, as in manual designed policies, is not an optimal strategy and introducing a diverse set of objects from other classes is beneficial.

\subsection{Automated Data Augmentation Benefits Large-Scale Data}\label{sec:waymo_result}
The Waymo Open Dataset is a recently released, large-scale dataset for 3D object detection in point clouds \cite{waymo2019}. The dataset contains roughly 20x more scenes than KITTI, and roughly 20x more human-annotated objects per scene. 
This dataset presents an opportunity to ask whether data augmentations -- being critical to model performance on the KITTI dataset due to the small size of the dataset -- continue to provide a benefit in a large-scale training setting more reflective of the self-driving conditions in the real world.

To address this question, we evaluate the proposed method on the Waymo Open Dataset. In particular, we evaluate PPBA with StarNet~\cite{ngiam2019starnet} and PointPillars~\cite{lang2019pointpillars} on the {\it test} split in Table~\ref{table:waymo_test_car} and Table~\ref{table:waymo_test_ped} on both LEVEL$\_1$ and LEVEL$\_2$ difficulties at different ranges. Our results indicates that PPBA notably improves the predictive accuracy of 3D detection across architectures, difficulty levels and object classes.
These results indicate that data augmentation remains an important method for boosting model performance even in large-scale dataset settings. Furthermore, the gains due to PPBA may be as large as changing the underlying architecture, without any increase in inference cost.

\begin{table}[h!]
\begin{center}
\caption{ Performance comparison on the Waymo Open Dataset {\it test} set for vehicle detection. Note that the results of PointPillars \cite{lang2019pointpillars} on the Waymo Open Dataset are reproduced by \cite{waymo2019}}
\label{table:waymo_test_car}
\resizebox{\textwidth}{!}{
\begin{tabular}{l||c|cccc|cccc}
\hline
\multirow{2}{*}{Method} & 
\multicolumn{1}{c|}{Difficulty} &
\multicolumn{4}{c|}{3D mAP (IoU=0.7)} & \multicolumn{4}{c}{3D mAPH  (IoU=0.7) } \\
& Level & Overall & 0-30m & 30-50m & 50m-Inf & Overall & 0-30m & 30-50m & 50m-Inf\\
\hline
StarNet \cite{ngiam2019starnet} & 1 & 61.5 & 82.2 & 56.6 & 32.2 & 61.0 & 81.7 & 56.0 & 31.8 \\
StarNet \cite{ngiam2019starnet} + PPBA & 1 & \bf64.6 & \bf85.8 & \bf59.5 & \bf35.1 & \bf64.1 & \bf85.3 & \bf58.9 & \bf34.6 \\
StarNet \cite{ngiam2019starnet} & 2 & 54.9 & 81.3 & 49.5 & 23.0 & 54.5 & 80.8 & 49.0 & 22.7 \\
StarNet \cite{ngiam2019starnet} + PPBA &  2 & \bf56.2 & \bf82.8 & \bf54.0 & \bf26.8 & \bf55.8 & \bf82.3 & \bf53.5 & \bf26.4 \\
\hline
PointPillars \cite{lang2019pointpillars} & 1 & 63.3 & 82.3 & 59.2 & 35.7 & 62.8 & 81.9 & 58.5 & 35.0 \\
PointPillars \cite{lang2019pointpillars} + PPBA &  1 & \bf67.5 & \bf86.7 & \bf63.5 & \bf39.4 & \bf67.0 & \bf86.2 & \bf62.9 & \bf38.7 \\
PointPillars \cite{lang2019pointpillars} & 2 & 55.6 & 81.2 & 52.9 & 27.2 & 55.1 & 80.8 & 52.3 & 26.7 \\
PointPillars \cite{lang2019pointpillars} + PPBA & 2 & \bf59.6 & \bf85.6 & \bf57.6 & \bf30.0 & \bf59.1 & \bf85.1 & \bf57.0 & \bf29.5 \\
\hline
\end{tabular}
}
\end{center}
\end{table}

\begin{table}[h!]
\begin{center}
\caption{ Performance comparison on the Waymo Open Dataset {\it test} set for pedestrian detection. Note that the results of PointPillars \cite{lang2019pointpillars} on the Waymo Open Dataset are reproduced by \cite{waymo2019}}
\label{table:waymo_test_ped}
\resizebox{\textwidth}{!}{
\begin{tabular}{l||c|cccc|cccc}
\hline
\multirow{2}{*}{Method} & 
\multicolumn{1}{c|}{Difficulty} &
\multicolumn{4}{c|}{3D mAP (IoU=0.5)} & \multicolumn{4}{c}{3D mAPH  (IoU=0.5) } \\
& Level & Overall & 0-30m & 30-50m & 50m-Inf & Overall & 0-30m & 30-50m & 50m-Inf\\
\hline
StarNet \cite{ngiam2019starnet} & 1 & 67.8 & 76.0 & 66.5 & 55.3 & 59.9 & 67.8 & 59.2 & 47.0 \\
StarNet \cite{ngiam2019starnet} + PPBA & 1 & \bf69.7 & \bf77.5 & \bf68.7 & \bf57.0 & \bf61.7 & \bf69.3 & \bf61.2 & \bf48.4 \\
StarNet \cite{ngiam2019starnet} & 2 & 61.1 & 73.1 & 61.2 & 44.5 & 54.0 & 65.2 & 54.5 & 37.8 \\
StarNet \cite{ngiam2019starnet} + PPBA & 2 & \bf63.0 & \bf74.8 & \bf63.2 & \bf46.5 & \bf55.8 & \bf66.8 & \bf56.2 & \bf39.4 \\
\hline
PointPillars \cite{lang2019pointpillars} & 1 & 62.1 & 71.3 & 60.1 & 47.0 & 50.2 & 59.0 & 48.3 & 35.8 \\
PointPillars \cite{lang2019pointpillars} + PPBA & 1 & \bf66.4 & \bf74.7 & \bf64.8 & \bf52.7 & \bf54.4 & \bf62.5 & \bf52.5 & \bf41.2 \\
PointPillars \cite{lang2019pointpillars} & 2 & 55.9 & 68.6 & 55.2 & 37.9 & 45.1 & 56.7 & 44.3 & 28.8 \\
PointPillars \cite{lang2019pointpillars} + PPBA & 2 & \bf60.1 & \bf72.2 & \bf59.7 & \bf42.8 & \bf49.2 & \bf60.4 & \bf48.2 & \bf33.4 \\
\hline
\end{tabular}
}
\end{center}
\end{table}

When performing the search with PPBA, 16 trials are trained to optimize the mAP for car and pedestrian, respectively. The list of augmentation operations described in Section~\ref{sec:search_space}, except for GroundTruthAugmentor and RandomFlip, are used during search. In our experiments, we have found RandomFlip has a negative impact on heading prediction for both car and pedestrian.

For both StarNet and PointPillars on the Waymo Open Dataset, the same training and inference settings\footnote{\url{http://github.com/tensorflow/lingvo}} as~\cite{ngiam2019starnet} is used. All trials are trained on the full {\it train} set and validated on the 10\% {\it validation} split (4,109 samples). For StarNet, we train the first iteration for 8,000 steps and the remaining iterations for 4,000 steps with batch size 128. The training steps for PointPillars are reduced by half in each iteration with batch size 64. We perform the search for 25 iterations on StarNet and for 20 iterations on PointPillars.

Even though StarNet and PointPillars are two distinct types of detection models, we have observed similar patterns in the evolution of their augmentation schedules. For StarNet and PointPillars, the probability of FrustumDropout is reduced from 100\% to 23\% and 56\%, and the maximum rotation angle in RandomRotation is reduced from 0.785 to 0.54 and 0.42. These examples indicate that applying weaker data augmentation towards the end of training is beneficial.

\subsection{Better Results with Less Computation}\label{sec:computation_efficiency}

Above, we have verified the effectiveness of PPBA on improving 3D object detection on the KITTI Dataset and the Waymo Open Dataset. In this section, we analyze the computational cost of PPBA, and compare PPBA with random search and PBA~\cite{ho2019population} on the KITTI {\it test} split.

All searches are performed with StarNet~\cite{ngiam2019starnet} and the search space described in Section~\ref{sec:search_space}. For Random Search\footnote{Our initial experiment on random search shows the performance distribution of augmentation policies is spread on the KITTI {\it validation} split. In order to save computation resources, the random search here is performed on a fine-grained search space.}, 1,000 distinct augmentation policies are randomly sampled and trained. PBA is run with 16 total trials while training the first iteration for 3,000 steps and the remaining iterations for 1,000 steps with batch size 64. 

The baseline StarNet is trained for 8 hours with a TPU v3-32 Pod~\cite{jouppi2017datacenter,tpuv3_2019} on vehicle detection and pedestrian/cyclist detection models. Random search requires about $1,000\times8 = 8,000$ TPU hours for training. In comparison, both PBA and PPBA train with a much smaller cost of $8\times16 = 128$ TPU hours, with an additional real-time computation overhead of waiting for the evaluation result for $8\times16 = 128$ TPU hours. We observe that PPBA results in a more than 30x speedup compared to random search, while identifying better-performing augmentation strategies. Furthermore, PPBA outperforms PBA by a substantial margin with the same computational budget.

\begin{table}
\small{
\begin{center}
\caption{Comparison of 3D mAP on StarNet on the KITTI {\it test} set across data augmentation methods}
\label{table:compare_search_algorithm}
\resizebox{\textwidth}{!}{
\begin{tabular}{l|c||ccc|ccc|ccc}
\hline
\multirow{2}{*}{Method} & 
\multirow{2}{*}{TPU Hours} & 
\multicolumn{3}{c|}{Car} & \multicolumn{3}{c|}{Pedestrian} & \multicolumn{3}{c}{Cyclist} \\
& & Easy & Mod. & Hard & Easy & Mod. & Hard & Easy & Mod. & Hard\\
\hline
Manual design~\cite{ngiam2019starnet}& 8 & 81.63 & 73.99 & 67.07 & 48.58 & 41.25 & 39.66 & 73.14 & 58.29 & 52.58 \\
Random Search & 8,000 & 81.89 & 74.94 & 67.39 & \bf52.78 & \bf44.71 & 41.12 & 73.71 & 59.92 & 54.09 \\
PBA & 256 & 83.16 & 75.02 & 69.72 & 41.28 & 34.48 & 32.24 & 76.8 & 59.43 & 52.77 \\
PPBA & 256 & \bf84.16 & \bf77.65 & \bf71.21 & 52.65 & 44.08 & \bf41.54 & \bf79.42 & \bf61.99 & \bf55.34 \\
\hline
\end{tabular}}
\end{center}
}
\end{table}

\begin{figure}[h]
    \centering
    \includegraphics[width=0.8\linewidth]{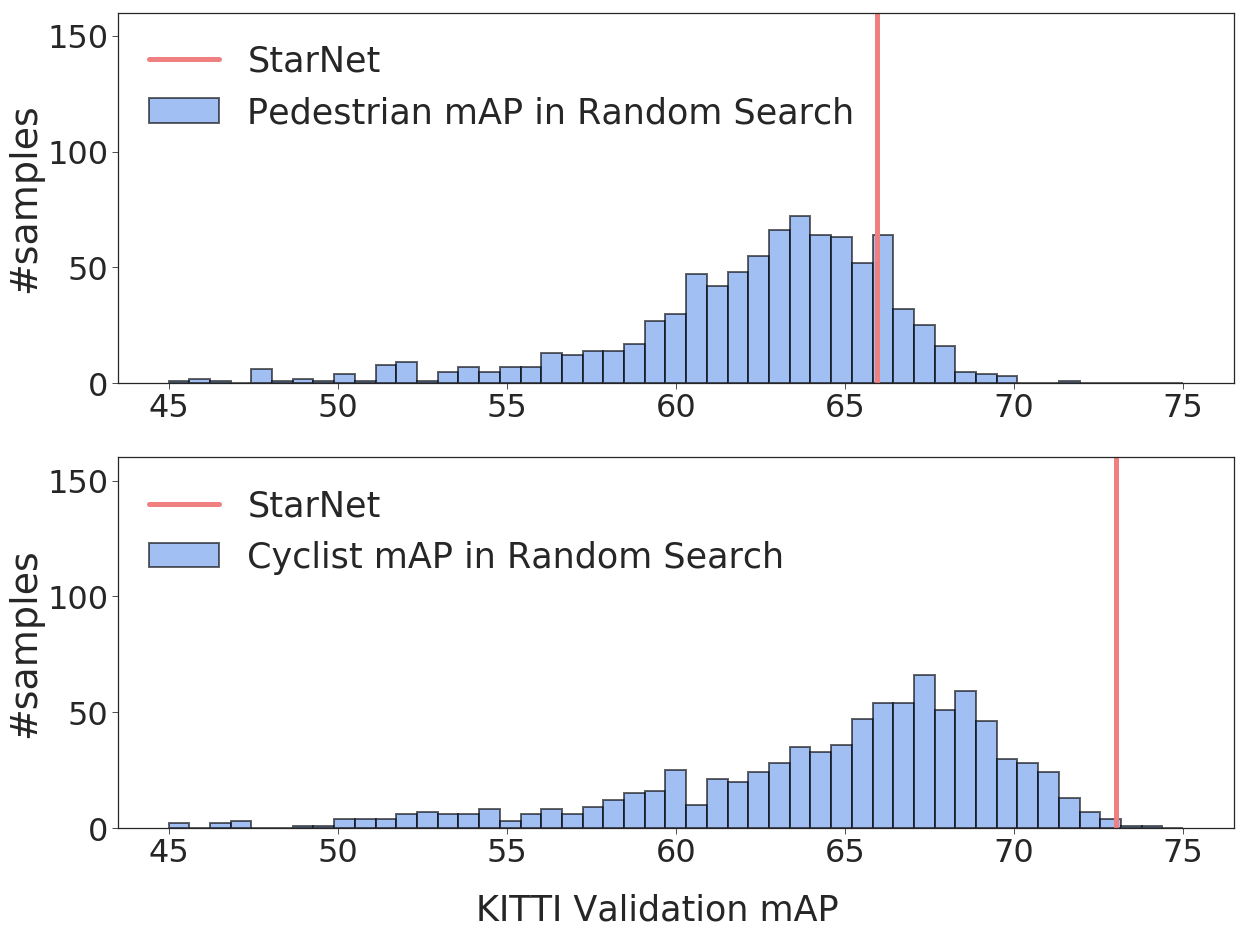}
    \caption{3D mAP of a population of 1,000 random augmentation policies for pedestrian and cyclist on the moderate difficulty on the KITTI {\it validation} split}
    \label{fig:baseline-vs-random-histogram}
\end{figure}

While searching the augmentation policies randomly for pedestrian/cyclist detection, the majority of samples perform worse than the manual designed augmentation strategy on the KITTI {\it validation} split (see Fig.~\ref{fig:baseline-vs-random-histogram}). Unlike image augmentation search spaces, where each operation has one parameter and even random policies lead to some improvement in generalization, point cloud augmentations are harder to optimize, with a larger number of parameters (e.g. geometric distance, operation strength, distribution of categorical sampling, etc.) and no good priors for the parameters' ranges. Because of the complex search space, it is challenging to discover good augmentation policies with random search, especially for the cyclist category. We find that it is effective to fine tune the parameter search space of each operation to improve the overall performance of random search. However, the whole process is expensive and requires domain expertise.

We observe that PBA is not effective at discovering better augmentation policies, compared to random search or even to manual search, when the detection category is sensitive to inferior augmentation parameters. In PBA, the full search space is explored at every iteration, which is inefficient for searching parameters in a high dimensional space. To mitigate the inefficiency, PPBA progressively explores a subset of search space at every iteration, and the best parameters discovered in past iterations are adopted in the exploration phase. As in Table~\ref{table:compare_search_algorithm}, PPBA shows much larger improvements on the car and cyclist categories, demonstrating the effectiveness of the proposed strategy.

\subsection{Automated Data Augmentation Improves Data Efficiency}\label{sec:data_efficiency}

In this section, we conduct experiments to determine how PPBA performs when the dataset size grows. The experiments are performed with PointPillars on subsets of the Waymo Open Dataset with the following number of training examples: 10\%, 30\%, 50\%, by randomly sampling run segments and single frames of sensor data, respectively. During training, the decay interval of the learning rate is linearly decreased accordingly to the percentile of data sampled (e.g., reduce the decay interval of learning rate by 50\% when sampling 50\% of the training examples), while the number of training epochs is set to be inversely proportional to the percentile of data sampled. As it is commonly known that smaller datasets need more regularization, we increase weight decay from $10^{-4}$ to $10^{-3}$, when training on 10\% of the examples.

\begin{table}[h!]
\begin{center}
\caption{Compare 3D mAP of PointPillars with no augmentation, random augmentation and PPBA on the Waymo Open Dataset {\it validation} set as the dataset size grows}
\label{table:waymo_reduced_dataset}
\resizebox{\textwidth}{!}{
\begin{tabular}{l|c||cc|cc|cc|cc}
\hline
\multirow{2}{*}{Method} & 
\multirow{2}{*}{Sample Unit} & 
\multicolumn{2}{c|}{10\%} & \multicolumn{2}{c|}{30\% } 
& \multicolumn{2}{c|}{50\% } & \multicolumn{2}{c}{100\% } \\

& & Car & Pedestrian & Car & Pedestrian & Car & Pedestrian & Car & Pedestrian\\
\hline
Baseline & run segment & 42.5 & 46.1 &49.5 & 56.4 &52.5 &59.1 & 57.2 &  62.3\\
Random & run segment & 49.5 & 50.6 & 54.1 & 58.8 & 56.1 & 60.5 & 60.9 & 63.5 \\
PPBA & run segment &\bf54.2 &\bf55.8 & \bf57.6 &\bf63.0 &\bf58.7 &\bf65.1 & \bf62.4 & \bf66.0 \\
\hline
Baseline & single frame & 52.4 & 56.9 & 55.3 & 60.7 & 56.7 & 61.2 & 57.2 & 62.3 \\
Random & single frame &58.3 & 59.8 & 59.4 & 61.9 & 59.7 & 62.1 & 60.9 & 63.5 \\
PPBA & single frame & \bf59.8 & \bf64.2 & \bf60.7 & \bf65.5 &\bf61.2 &\bf66.2 & \bf62.4 & \bf66.0 \\

\hline
\end{tabular}
}
\end{center}
\end{table}

\begin{figure}[h!]
    \centering
    \includegraphics[width=1\linewidth]{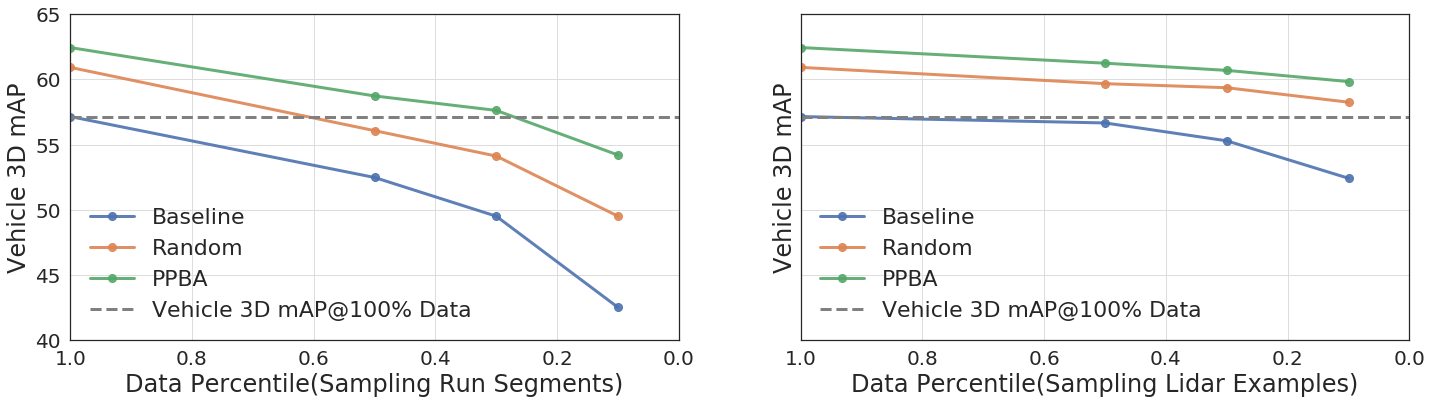}
    \caption{Vehicle detection 3D mAP for PointPillars on Waymo Open Dataset {\it validation} set with no augmentation, random augmentation and PPBA as the dataset size changes}
    \label{fig:reduced_data_car}
\end{figure}

\begin{figure}[h!]
    \centering
    \includegraphics[width=1\linewidth]{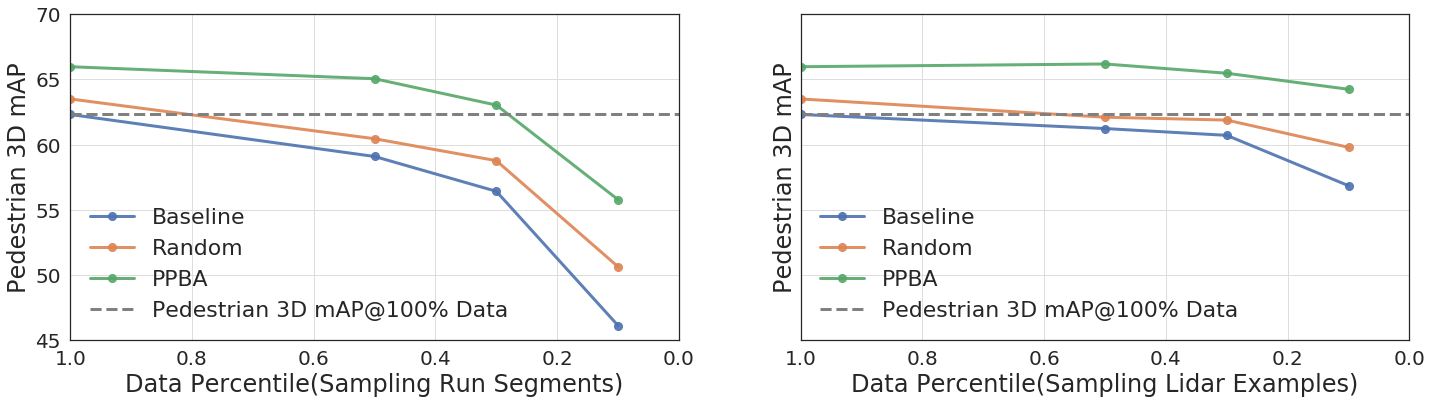}
    \caption{Pedestrian detection 3D mAP for PointPillars on Waymo Open Dataset {\it validation} set with no augmentation, random augmentation and PPBA as the dataset size changes}
    \label{fig:reduced_data_ped}
\end{figure}

Compared to downsampling from single frames of sensor data, performance degradation of PointPillars models is more severe when downsampling from run segments. This phenomenon is due to the relative lack of diversity in the run segments, which tend to contain the same set of distinct vehicles and pedestrians. As in Table~\ref{table:waymo_reduced_dataset}, Fig.~\ref{fig:reduced_data_car} and Fig.~\ref{fig:reduced_data_ped}, we compare the overall 3D detection mAP on the Waymo Open Dataset {\it validation} set for all ground truth examples with $\geq$ 5 points and rated as LEVEL$\_1$ difficulty for 3 sets of PointPillars models: with no augmentation, random augmentation policy and PPBA. While random augmentation policy can improve the PointPillars baselines and demonstrate the effectiveness of the proposed search space, PPBA pushes the limit even further. PPBA is 10x more data efficient when sampling from single frames of sensor data, and 3.3x more data efficient when sampling from run segments. As we expected, the improvement from PPBA becomes larger when the dataset size is reduced.

\subsection{Progressive Population Based Augmentation Generalizes on Image Classification}\label{sec:ppba_generalization}

So far our experiments have demonstrated that PPBA consistently improves over alternatives for 3D object detection across datasets and architectures. However, PPBA is a general algorithm, and in this section we validate its true versatility by applying it to a common 2D image classification problem.

To search for augmentation policies, we use the same reduced subset of the ImageNet training set with 120 classes and 6,000 samples as~\cite{cubuk2018autoaugment,lim2019fast}. During the PPBA search, 16 trials are trained to optimize the Top-1 accuracy on the reduced validation set for 8 iterations while 4 operations are selected for exploration at every iteration. When replaying the learned augmentation schedule on the full training set, the ResNet-50 model is trained for 180 epochs with a batch size of 4096, a weight decay of $10^{-4}$ and a cosine decay learning rate schedule with learning rate of 1.6. The results on the ImageNet validation set, shown in Table~\ref{table:resnet50_imagenet}, confirm that PPBA can be used as a highly efficient auto augmentation algorithm for tasks other than 3D object detection.

\begin{table}[h!]
\begin{center}
\caption{Comparison of Top-1 accuracy (\%) and computational cost across augmentation methods on the ImageNet validation set for ResNet-50. Note that the baseline results with Inception-style Pre-processing is reproduced by~\cite{cubuk2018autoaugment}}
\label{table:resnet50_imagenet}
\begin{tabular}{l|c|c|c}
\hline
Method & Accuracy & GPU Hours & Hardware\\
\hline
Inception-style Pre-processing~\cite{szegedy2015going} & 76.3  & - & - \\
AutoAugment~\cite{cubuk2018autoaugment} &77.6&15000 & GPU P100 \\
Fast AutoAugment~\cite{lim2019fast} &77.6 &450 & GPU V100\\
PPBA &77.5 & 16 & GPU V100\\
\hline
\end{tabular}
\end{center}
\end{table}
\section{Conclusion}
We have presented Progressive Population Based Augmentation, a novel automated augmentation algorithm for point clouds. PPBA optimizes the augmentation schedule via narrowing down the search space and adopting the best parameters from past iterations. Compared with random search and PBA, PPBA can more effectively and more efficiently discover good augmentation policies in a rich search space for 3D object detection. Experimental results on the KITTI dataset and the Waymo Open Dataset demonstrate that the proposed method can significantly improve 3D object detection in terms of performance and data efficiency. While we have also validated the effectiveness of PPBA on a common task such as image classification, exploring the potential applications of the algorithm to more tasks and models remains an exciting direction of future work.

\clearpage
%
%
\bibliographystyle{splncs04}
\bibliography{refs}

\clearpage

\appendix
\section{Supplementary materials for ``Improving 3D Object Detection through Progressive Population Based Augmentation''}\label{sec:appendix}

\begin{center}
\captionof{table}{List of point cloud transformations in the search space for point cloud 3D object detection}
\label{table:operations1}\par
\resizebox{\textwidth}{!}{
\begin{tabular}{l p{10cm}}
\hline
Operation Name & Description\\
\hline
GroundTruthAugmentor~\cite{yan2018second} & Augment the bounding boxes from a ground truth data base ($<25$ boxes per scene)\\
RandomFlip~\cite{yang2018pixor} & Randomly flip all points along the Y axis.\\
WorldScaling~\cite{zhou2018voxelnet} & Apply global scaling to all ground truth boxes and all points.\\
RandomRotation~\cite{zhou2018voxelnet} & Apply random rotation to all ground truth boxes and all points.\\
GlobalTranslateNoise & Apply global translating to all ground truth boxes and all points along x/y/z axis.\\
FrustumDropout & All points are first converted to spherical coordinates, and then a point is randomly selected. All points in the frustum around that point within a given phi, theta angle width and distance to the
original greater than a given value are dropped randomly.\\
FrustumNoise & Randomly add noise to points within a frustum in a converted spherical coordinates.\\
RandomDropout & Randomly dropout all points.\\
\hline
\end{tabular}
}
\end{center}

\smallskip

\begin{center}
\captionof{table}{The range of augmentation parameters that can be searched by Progressive Population Based Augmentation algorithm for each operation}
\label{table:operations2}\par
\vspace{-0.3cm}
\resizebox{\textwidth}{!}{
\begin{tabular}{l p{7cm} l}
\hline
Operation Name & Parameter Name & Range \\
\hline
\multirow{4}{*}{GroundTruthAugmentor} & vehicle sampling probability  & [0, 1] \\
& pedestrian sampling probability & [0, 1] \\
& cyclist sampling probability & [0, 1] \\
& other categories sampling probability & [0, 1] \\
\hline
RandomFlip & flip probability & [0, 1] \\
\hline
WorldScaling & scaling range & [0.5, 1.5] \\
\hline
RandomRotation & maximum rotation angle & [0, $\pi/4$] \\
\hline
\multirow{3}{*}{GlobalTranslateNoise} & standard deviation of noise on x axis  & [0, 0.3]\\
& standard deviation of noise on y axis & [0, 0.3]\\
& standard deviation of noise on z axis & [0, 0.3]\\
\hline
\multirow{5}{*}{FrustumDropout} & theta angle width of the selected frustum & [0, 0.4]\\
& phi angle width of the selected frustum & [0, 1.3] \\
& distance to the selected point & [0, 50] \\
& the probability of dropping a point & [0, 1] \\
& drop type\footnotemark & \{'union', 'intersection'\} \\
\hline
\multirow{5}{*}{FrustumNoise} & theta angle width of the selected frustum & [0, 0.4] \\
& phi angle width of the selected frustum & [0, 1.3] \\
& distance to the selected point & [0, 50] \\
& maximum noise level & [0, 1] \\
& noise type\footnotemark & \{'union', 'intersection'\} \\
\hline
RandomDropout & dropout probability & [0, 1] \\
\hline
\end{tabular}}
\end{center}

\addtocounter{footnote}{-2}
\stepcounter{footnote}\footnotetext{\fontsize{7}{7} \selectfont{Drop points in either the union or intersection of phi width and theta width.}}
\stepcounter{footnote}\footnotetext{\fontsize{7}{7} \selectfont{Add noise to either the union or intersection of phi width and theta width.}}

\begin{algorithm}[h!]
\caption{Progressive Population Based Augmentation}\label{alg:PPBA}
\begin{algorithmic}
\STATE {\bf Input}: data and label pairs $(\mathcal{X}, \mathcal{Y})$
\STATE {\bf Search Space}: $\mathcal{S} = \{op_i: params_i\}_{i=1}^n$
\STATE Set $t = 0$, $num\_ops = 2$, population $\mathcal{P} = \{\}$, best params and metrics for each operation $historical\_op\_params = \{\}$
\WHILE{$t \neq \mathcal{N}$}
\FOR{$\theta_i^t$ in $\{\theta_1^t, \theta_2^t, ..., \theta_\mathcal{M}^t\}$ (asynchronously in parallel)}
\STATE {\bf \# Initialize models and augmentation parameters in current iteration}
\IF {$t == 0$}
\STATE $op\_params_i^t$ = Random.sample($\mathcal{S}$, $num\_ops$)
\STATE Initialize $\theta_i^t$, $\lambda_i^t$, $params$ of $op\_params_i^t$
\STATE Update $\lambda_i^t$ with $op\_params_i^t$
\ELSE
\STATE Initialize $\theta_i^t$ with the weights of $winner_i^{t-1}$
\STATE Update $\lambda_i^t$ with $\lambda_i^{t-1}$ and $op\_params_i^t$
\ENDIF
\STATE {\bf \# Train and evaluate models, and update the population}
\STATE Update $\theta_i^t$ according to formular~\eqref{eq:2}
\STATE Compute metric $\Omega_i^t = \Omega(\theta_i^t)$
\STATE Update $historical\_op\_params$ with $op\_params_i^t$ and $\Omega_i^t$
\STATE $\mathcal{P} \leftarrow \mathcal{P} \cup \{\theta_i^t\}$
\STATE {\bf \# Replace inferior augmentation parameters with better ones}
\STATE $winner_i^t \leftarrow$  Compete($\theta_i^t$, Random.sample($\mathcal{P}$))
\IF{$winner_i^t \neq \theta_i^t$}
\STATE $op\_params_i^{t+1}  \leftarrow$ Mutate($winner_i^t$'s $op\_params$, $historical\_op\_params$)
\ELSE
\STATE $op\_params_i^{t+1} \leftarrow op\_params_i^{t}$
\ENDIF
\ENDFOR
\STATE $t \leftarrow t + 1$
\ENDWHILE
\end{algorithmic}
\end{algorithm}

\begin{algorithm}[h!]
\caption{Exploration Based on Historical Data}\label{alg:Exploration}
\begin{algorithmic}
\STATE {\bf Input}: $op\_params = \{op_i: params_i\}_{i=1}^{num\_ops}$, best params and metric for each operation $historical\_op\_params$
\STATE {\bf Search Space}: $\mathcal{S} = \{(op_i, params_i)\}_{i=1}^n$
\STATE Set $exploration\_rate = 0.8$, $selected\_ops = []$, $new\_op\_params = \{\}$
\IF{Random(0, 1) $< exploration\_rate$}
\STATE $selected\_ops$ = $op\_params$.Keys()
\ELSE
\STATE $selected\_ops$ = Random.sample($\mathcal{S}$.Key(), $num\_ops$)
\ENDIF
\FOR{i in Range($num\_ops$)}
\STATE {\bf \# Choose augmentation parameters, which successors will mutate}
\STATE {\bf \# to generate new parameters}
\IF{$selected\_ops[i]$ in $op\_params$.Keys()}
\STATE $parent\_params = op\_params[selected\_ops[i]]$
\ELSIF{$selected\_ops[i]$ in $historical\_op\_params$.Keys()}
\STATE $parent\_params = historical\_op\_params[selected\_ops[i]]$
\ELSE
\STATE Initialize $parent\_params$ randomly
\ENDIF
\STATE $new\_op\_params[selected\_ops[i]] =$ MutateParams($parent\_params$)
\ENDFOR
\end{algorithmic}
\end{algorithm}

\clearpage

\section*{Acknowledgements}
We would like to thank Peisheng Li, Chen Wu, Ming Ji, Weiyue Wang, Zhinan Xu, James Guo, Shirley Chung, Yukai Liu, Pei Sun of Waymo and Ang Li of DeepMind for helpful feedback and discussions. We also thank the larger Google Brain team including Matthieu Devin, Zhifeng Chen, Wei Han and Brandon Yang for their support and comments.

\end{document}